%% file: main.tex
\documentclass[10pt,twocolumn,letterpaper]{article}

\usepackage{cvpr}              


\definecolor{cvprblue}{rgb}{0.21,0.49,0.74}
\usepackage[pagebackref,breaklinks,colorlinks,allcolors=cvprblue]{hyperref}

\newcommand{\hyper}[1]{\textcolor{magenta}{\texttt{#1}}}

\def\paperID{*****} 
\def\confName{CVPR}
\def\confYear{2025}

\title{Improving Keystep Recognition in Ego-Video via Dexterous Focus}

\author{Zach Chavis \quad Stephen J. Guy \quad Hyun Soo Park \vspace{0.3em} \\
{\normalsize University of Minnesota}\\
{\tt\small\hyper{https://appliedmotionlab.github.io/dexfocus}}\\
\vspace{-10mm}
}

\begin{document}
\maketitle

\begin{abstract}
    In this paper, we address the challenge of understanding human activities from an egocentric perspective. Traditional activity recognition techniques face unique challenges in egocentric videos due to the highly dynamic nature of the head during many activities. We propose a framework that seeks to address these challenges in a way that is independent of network architecture by restricting the ego-video input to a stabilized, hand-focused video. We demonstrate that this straightforward video transformation alone outperforms existing egocentric video baselines on the Ego-Exo4D Fine-Grained Keystep Recognition benchmark~\cite{egoexo4d} without requiring any alteration of the underlying model infrastructure.
\end{abstract}

\section{Introduction}

Understanding human motion from video is an important and well-studied area of computer vision \cite{humanactionsurvey}.
With the rise of all-day wearable AR smart glasses, egocentric video (ego-video) data has become increasingly studied~\cite{aria}, leading to the release of large ego-video datasets of human activities ~\cite{hot3d, Damen2022RESCALING, ego4d, egoexo4d} helping to drive research in areas of human activity understanding for personal AI assistants. However, ego-video activity analysis still lags behind exo-video benchmarks~\cite{goyal2017something,internvideo,DSCNet}, due to ego-video's highly unstable and dynamic nature, misalignment between view direction and the camera wearer's attention and intention, and limited visible context.

To overcome these challenges, we explore the idea of dexterous video focusing, where the egocentric video is cropped and stabilized to produce a hand-focused video.
Hand-focused video has several advantages over the raw ego-video. For many tasks, hands, and their surrounding context, convey reliable information about a person's current actions and proficiency. The context provided by hand-focused videos can range from the fine-manipulation of objects, to large sweeping motions present in physical tasks. Further, the hands can provide additional stability within the video -- while egocentric motion is often highly dynamic resulting in large context shifts frame to frame, hands provide an anchor to the scene, allowing for relevant information to move to and from the hands.

The contributions of this work are as follows:
\begin{itemize}
    \item A framework for extracting a stabilized, hand-focused video from egocentric videos.
    \item Showcasing that switching from full-frame egocentric video to hand-focused video is sufficient to achieve a broad-base improvement on keystep recognition with no alteration of the underlying model infrastructure.
    \item Combining full egocentric video and the respective hand-focused video boosts performance beyond each type of video alone, achieving substantial improvements on fine-grained keystep recognition.
\end{itemize}

\begin{figure}[t]
  \centering
   \includegraphics[width=0.98\linewidth]{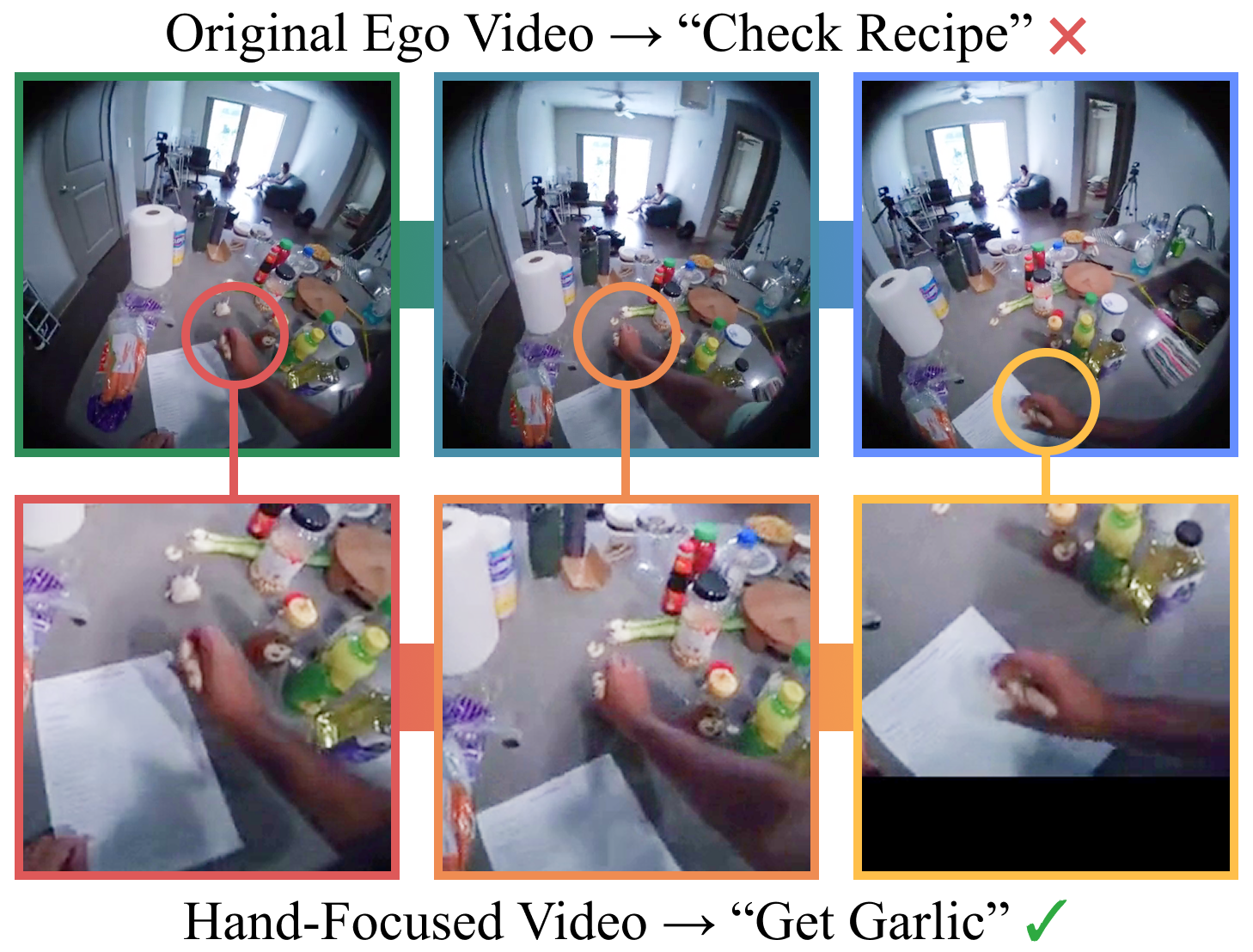}

   \caption{\textbf{Dexterous Focus.} Egocentric videos often contain significant dynamic motion, head tilting, and distracting elements in e.g. dexterous tasks. By restricting the ego video to only tracking the area around the camera-wearer's hands, we allow the model to mainly focus on relevant features for activity understanding, and we show improved performance on human activity understanding without needing to augment existing video network architectures.}
   \label{fig:teaser}
\end{figure}

\section{Related Work} 
\paragraph{Egocentric Activity Datasets.} The availability of large-scale third-person activity datasets~\cite{kinetics,goyal2017something,thumos14,activitynet} has enabled significant progress in action recognition and other video understanding tasks. To support ego-video activity analysis large-scale egocentric datasets are beginning to emerge, including Ego4D~\cite{ego4d}, Ego-Exo4D~\cite{egoexo4d}, EPIC-Kitchens~\cite{Damen2022RESCALING},  and HOT3D~\cite{hot3d}, which provide a variety of participants, scenarios, modalities, and activities.

\paragraph{Egocentric Video Analysis} General purpose, end-to-end video analysis models such as TimeSformer~\cite{timesformer} have been shown to be generally applicable to egocentric video~\cite{egoexo4d}. Specialized egocentric approaches have been developed, such as leveraging large datasets to learn rich egocentric features, as seen in EgoVLP~\cite{egovlp}, EgoVLPv2~\cite{egovlpv2}, and EgoVideo~\cite{egovideo}. These egocentric video features can be used in downstream models such as ActionFormer~\cite{actionformer} to improve performance on egocentric videos \cite{actionformerego4d}. An alternative approach to learning new encoders is presented in X-MIC~\cite{xmic}, which takes existing video encoding models trained on third-person data and learns a model to align the exocentric representation space using egocentric features such as hands. In situations where multi-modal data is available, incorporating this additional data can improve egocentric video performance~\cite{transfermae, egoexo4d, epicfusion, mmix, egoenv}.

\section{Dexterous Focus Approach}

We focus on extracting a stable, hand-focused video $\mathcal{V}_{\text{hands}}$ from an egocentric video $\mathcal{V}_{\text{ego}}$, which can be directly substituted for the input in modern video architectures~\cite{timesformer,actionformer}.

\paragraph{Dexterous Focus}
We first use the 100DOH~\cite{handcrop} hand detector $D$, which has been trained on ego-images to detect bounding boxes for all hands in a frame. Next, we define a function $F$ which selects the frame's focal point from the detected hand bounding boxes. Due to the possible presence of multiple people in the video, we filter out small-sized and low-confidence hand detections to isolate only the camera-wearer's visible hand(s). We then compute a single position per-frame $\mathbf{x}_{\text{hands}}$ representing the centroid of the camera-wearer's hand(s). In the event that the hand detector fails or the camera wearer's hands are not visible, we choose a fallback position at the bottom center of the frame, which is the average hand location in the dataset. 

Composing these functions over all frames results in the following trajectory:

\begin{equation}
\mathcal{X}_{\text{hands}} = F \circ D(\mathcal{V}_{\text{ego}}).
\end{equation}

\paragraph{Trajectory Stabilization} To mitigate the temporal noise induced by the per-frame functions $D$ and $F$, we apply a post-process smoothing kernel $\mathcal{S}$ to the hands' trajectory:
\begin{equation}
\tilde{\mathcal{X}}_{\text{hands}} = \mathcal{S}(\mathcal{X}_{\text{hands}}).
\end{equation}

\paragraph{Rendering} Finally, we crop the original video to the smoothed hand positions for every frame, enabling the downstream model to focus on hand context and dexterous activity. The final hand-focused video is defined as follows:

\begin{equation}
\mathcal{V}_{\text{hands}} = \operatorname{crop}(\mathcal{V}_{\text{ego}},\; \tilde{\mathcal{X}}_{\text{hands}}).
\end{equation}

\section{Activity Recognition}
To evaluate the effect of our dexterous focus framework on action recognition, we apply it to the Fine-Grained Keystep Recognition benchmark presented in Ego-Exo4D~\cite{egoexo4d}. The benchmark requires training a network that can recognize the current step a user is performing in a multi-step procedural task such as cooking or bike repair all from a single egocentric video clip. Much of the difficulty in this benchmark stems from different keysteps belonging to the same task looking very similar from the egocentric perspective, such as grabbing two similar but different items or rotating a dial clockwise or counterclockwise. We hypothesize that hand-focused videos provide an opportunity for the network to focus on these key details, allowing for improved performance.

\subsection{Benchmark \& Experimental Setup}
Video clips expressing keysteps in the Ego-Exo4D Fine-Grained Keystep Recognition benchmark are represented over different time spans, ranging from under one second (grabbing the garlic) to over five minutes (changing a bike wheel). The input to the network is the untrimmed video clip, and the output is a keystep-class prediction from 278 classes representing different specific steps across three types of procedural tasks (Cooking, Health Testing, and Bike Repair).

For this task, we re-implement the TimeSformer architecture baseline, which uses space-time attention to classify short video clips~\cite{timesformer}. We sample eight frames from the 448p video, with a delta-time between frames of $1.07$ seconds ($32$ frames). Due to the variable video length presented in this task, extremely short clips exceed this delta, therefore in our implementation we warp the delta-time, such that clips under $8\times32$ frames are sampled with a $dt=n/8$, where $n$ is the number of frames in the keystep clip. This effectively ``speeds-up" short clips, but allows the model to see more information. Similarly to the original TimeSformer baseline, we initialize our model with weights pretrained on the third-person action dataset Kinetics-600~\cite{kinetics}. To satisfy the pre-trained model's input dimensions, $V_{\text{ego}}$ is down-sampled to 224p, while $V_{\text{hands}}$ uses a square crop of 25\%, down to 224p. To combine both $V_{\text{ego}}$ and $V_{\text{hands}}$ video streams, we use a late-fusion strategy, employing dual TimeSformers, one for each video stream, and concatenating the encodings before passing to the classification head. We train across four V100 GPUs for 50 epochs, and report the model which maximizes validation accuracy.

\subsection{Results}

When looking at only hands, we see a $17\%$ improvement over full ego. This suggests that for keystep recognition tasks, hands provide a stable signal with sufficient relevant context. When combining both the full ego and hand streams, we reach $47.75\%$ accuracy, a $22\%$ improvement over ego-alone (Table~\ref{tab:keystep}).

\begin{table}[ht]
\centering
\begin{tabular}{lc|c}
   \toprule
      Method (pretraining) &  Train data &  Acc. (\%) \\ 
   \midrule
    TimeSFormer (K600)   & ego & 39.18\\
    TimeSFormer (K600)   & hands & 45.81 (+17\%)\\
    TimeSFormers (K600)   & ego+hands & 47.75 (+22\%)\\
   \bottomrule
\end{tabular} \\
\makeatletter
\def\@captype{table}
\makeatother
\caption{\small{\textbf{Keystep Recognition with Hands.} The Top-1 Accuracy of keystep recognition on hold-out validation dataset. We see that focusing on hands results in a $17\%$ improvement over ego alone, and both combined results in a $22\%$ improvement.}}
\label{tab:keystep}
\end{table}

\subsubsection{Benchmark Performance}

When comparing to the results presented in Ego-Exo4D, we find that our re-implementation of TimeSformer sees a $12\%$ improvement on Top-1 accuracy over its equivalent baseline, with no difference in the pretrained weights or training data. We suspect this is due to our time-warping, as a majority of clips in the benchmark are very short. 

As compared to existing state of the art as reported in the EgoExo4D benchmark~\cite{egoexo4d}, using our framework with only hands provides a $14\%$ improvement over the previous highest performing model which pretrained on both Ego and Exo viewpoints to learn a View-Invariant encoding~\cite{contrastive}. This performance imporve comes inspite of training on only a single mode of data with no acess to the exo view at training time. When using both ego+hands our approach has an $18\%$ improvement over state-of-the-art (Table~\ref{tab:keystepall}).

We expect there may be similar improvements by incorporating dexterous focus in other techniques such as Viewpoint Distillation and VI Encoder, but memory limitations may provide additional challenges at training time.

\begin{table}[ht]
\centering
\begin{tabular}{lc|c}
   \toprule
      Method (pretraining) &  Train data &  Acc. (\%) \\ 
   \midrule
    TimeSFormer (K600)   & exo & 32.68\\
    TimeSFormer (K600)   & ego & 35.13\\
    EgoVLPv2 (Ego4D)   & ego,exo & 35.84\\
    EgoVLPv2 (EgoExo4D)   & ego & 36.04\\
    EgoVLPv2 (Ego4D)   & ego & 36.51\\
    Ego-Exo Transfer MAE & ego,exo & 37.17\\
    Viewpoint Distillation   & ego,exo & 38.19\\
    EgoVLPv2 (EgoExo4D)   & ego,exo & 39.10\\
    TimeSFormer\text{*} (K600)   & ego & 39.18\\
    VI Encoder (EgoExo4D)   & ego,exo & 40.34\\
    TimeSFormer\text{*} (K600)   & hands & \underline{45.81}\\
    TimeSFormers\text{*} (K600)   & ego+hands & \textbf{47.75}\\
   \bottomrule
\end{tabular} \\
\makeatletter
\def\@captype{table}
\makeatother
\caption{\small{\textbf{Keystep Recognition Benchmark.} The Top-1 Accuracy of keystep recognition on hold-out validation dataset. Star (\text{*}) denotes our TimeSFormer re-implementation, with all other results reported directly from Ego-Exo4D~\cite{egoexo4d}. Rows are ranked by performance.}}
\label{tab:keystepall}
\end{table}

\section{Discussion}
Our results show that using smoothed hand locations to provide a visual focus to egocentric videos can have a significant impact on the accuracy of keystep classification. 
While we saw the best performance combining hand-focused video with the original input, the hand-focused video alone provided a broad-based improvement and should be considered when designing future egocentric video analysis frameworks.

\textbf{Limitations.} During certain tasks, hands may seldom be visible from an ego-view (e.g., combing one's hair, or moving furniture), especially with smaller camera FOVs. Because our method preprocesses the dataset, the hyperparameters are fixed across all scenarios.

\textbf{Future Work.} One important step for future work is considering ways to reduce memory requirements when operating on both ego and hand-focused video. Selective attention between corresponding parts of video \cite{xmic} or context summarizing vectors \cite{egoenv} are promising options. While the hand has been proven to be a powerful signal, there may be other signals (e.g., local motion metrics, other people, spoken instructions, gaze direction) to guide focus, and a general focus guiding network would be the most general option. In datasets with exocentric cameras, these exterior views may provide additional important context that could drive video focus. Finally, we are especially interested in future multi-modal applications where additional sensors commonly found in wearable technology (IMUs, audio data, etc.) can be integrated with hand-focus techniques to improve overall action recognition.

{
    \small
    \bibliographystyle{ieeenat_fullname}
    \bibliography{main}
}

\end{document}